\documentclass{article}

\PassOptionsToPackage{numbers,compress}{natbib}
\usepackage[preprint]{neurips2026/neurips_2026}
\makeatletter
\renewcommand{\@noticestring}{Preprint. Under review.}
\makeatother

\usepackage{amsmath,amsfonts,bm}

\def\eqref#1{equation~\ref{#1}}

\def\1{\bm{1}}

\DeclareMathAlphabet{\mathsfit}{\encodingdefault}{\sfdefault}{m}{sl}
\SetMathAlphabet{\mathsfit}{bold}{\encodingdefault}{\sfdefault}{bx}{n}

\usepackage{pgfplots}
\pgfplotsset{compat=1.18}
\usepackage{tikz}
\usepackage{multirow}
\usepackage[many]{tcolorbox}
\usepackage{wrapfig}
\usepackage{amsmath}
\usepackage{amsthm}
\usepackage{amssymb}
\usepackage{mathtools}
\usepackage{microtype}
\usepackage{graphicx}
\usepackage{float}
\usepackage{booktabs}
\usepackage{tabularx}
\usepackage{threeparttable}
\usepackage{siunitx}
\usepackage{colortbl}
\usepackage{xspace}
\usepackage[utf8]{inputenc}
\usepackage[T1]{fontenc}
\usepackage{hyperref}
\usepackage{url}
\usepackage{nicefrac}
\usepackage{xcolor}
\usepackage{subcaption}
\usepackage{enumitem}
\usepackage[capitalize,noabbrev]{cleveref}

\graphicspath{{./}{figures/}{neurips2026/}}

\newtheorem{proposition}{Proposition}

\title{BoHA: Blockwise Hadamard Product Adaptation\\
       for Parameter-Efficient Fine-Tuning}

\author{%
  Feng Yu\thanks{Correspondence: \texttt{fy274@exeter.ac.uk}}\quad
  Jia Hu\quad
  Geyong Min\\[0.3em]
  Department of Computer Science\\
  University of Exeter
}

\begin{document}

\maketitle

\begin{abstract}
Parameter-efficient fine-tuning (PEFT) of large language models trains
a small task-specific parameter set while keeping the pretrained
model frozen. The dominant Low-Rank Adaptation (LoRA) family makes
this trade-off practical; however, evaluations under the same parameter
budget assess single-task accuracy. In sequential adaptation settings,
such evaluations should also measure how well performance on the
first-stage task is retained after subsequent fine-tuning.
To address this gap, we introduce BoHA, a blockwise $W_0$-coupled
Hadamard product adapter that treats spatial support as an explicit
design axis. BoHA partitions the frozen weight $W_0$ into a
$b{\times}b$ grid and learns an independent low-rank Hadamard product
factor in each block, preserving a matched LoRA-equivalent total rank
with adapter-free merged inference. On a synthetic target, BoHA at
per-block rank $r_b{=}1$ exactly reconstructs an update that requires rank $b^2$
under the global $W_0$-coupled Hadamard parameterization. Across
Llama-3.2-1B/3B, Mistral-7B, and Gemma-2-9B on commonsense and
arithmetic reasoning tasks, BoHA outperforms LoRA across all
matched-budget single-task averages and remains competitive with the
strongest Hadamard baseline. On a Llama-3.2-3B
commonsense\,$\to$\,arithmetic continual-learning diagnostic, BoHA
retains $57.66\%$ first-stage accuracy and exceeds the $W_0$-free
additive-control mean by $15.23\%$ under matched second-stage
plasticity. These results demonstrate that blockwise $W_0$-coupled
Hadamard adaptation is a competitive PEFT design choice when retention
under sequential adaptation is part of the objective.
\end{abstract}

\section{Introduction}
\label{sec:intro}

Adapting large language models by full fine-tuning is costly when one
base model must serve many specialized
deployments~\citep{brown2020gpt3,devlin2019bert}. Parameter-efficient
fine-tuning (PEFT) reduces this cost by training a small task-specific
parameter set while freezing the pretrained
model~\citep{houlsby2019adapter,hu2022lora}. The dominant
parameterization is Low-Rank Adaptation (LoRA)~\citep{hu2022lora},
which adds a low-rank trainable update $\Delta W$ to the frozen
weight $W_0$, motivated by the finding that fine-tuning trajectories
lie in low-dimensional subspaces~\citep{aghajanyan2021intrinsic}. Subsequent PEFT research
refines this parameterization along several axes: reweighting or
reinitializing the additive
update~\citep{liu2024dora,meng2024pissa}, relaxing LoRA's rank
constraint by composing the update multiplicatively with
$W_0$~\citep{huang2025hira} or via a Hadamard product without
coupling to $W_0$~\citep{singhal2026abba}, or imposing block or
Kronecker structure on the additive
factors~\citep{jung2025gralora,edalati2022krona,yeh2024lokr}.

The taxonomy above varies how the additive update is reweighted,
decomposed, or block-tiled, while multiplicative coupling to $W_0$ is
either fixed in a single global Hadamard form~\citep{huang2025hira}
or removed entirely, leaving the update $W_0$-free. These designs are
usually compared by fixing trainable parameter count and reporting
single-task accuracy. Under sequential adaptation, a later
fine-tuning stage can degrade earlier-task
performance~\citep{mccloskey1989catastrophic,kirkpatrick2017ewc,delange2022continual},
which matched-budget single-task evaluations do not capture. This
motivates a research question: \emph{how should an adapter allocate
its update across a weight matrix when both matched-budget accuracy
and retained accuracy after subsequent fine-tuning matter?}

\begin{figure}[t]
\centering
\includegraphics[width=0.85\linewidth]{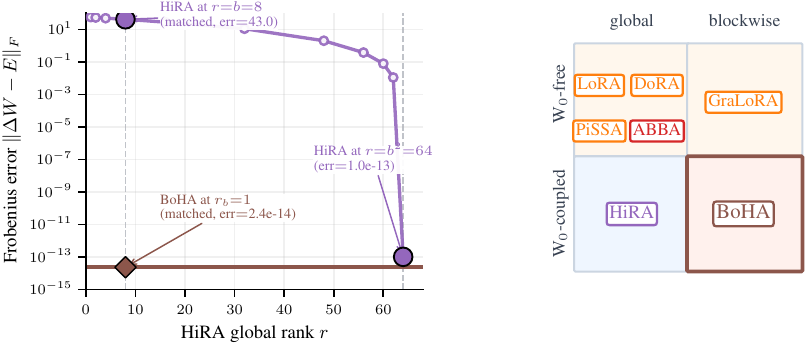}
\caption{\textbf{BoHA synthetic construction and design axes.}
(a) On a synthetic target where $E \oslash W_0$ has per-block rank
$1$ and global rank $b^2$, BoHA reconstructs at $r_b{=}1$ while a
global Hadamard parameterization needs global rank $b^2$.
(b) PEFT methods organized by $W_0$-coupling and spatial support.
BoHA occupies the blockwise $W_0$-coupled cell.}
\label{fig:hero}
\end{figure}

We answer this question by separating two structural axes the
single-task protocol leaves implicit: $W_0$-coupling (whether
$\Delta W$ is multiplicatively coupled to the frozen weight at
evaluation) and spatial support (global vs.\ blockwise). We introduce
Blockwise Hadamard Product Adaptation (BoHA), which combines blockwise
support with $W_0$-coupled Hadamard modulation: a $b{\times}b$
partition of $W_0$
with an independent low-rank Hadamard product per block, at
LoRA-equivalent total rank, as shown in \autoref{fig:hero}(b). The analysis isolates what blockwise localization changes: with $E$
denoting the full fine-tuning update, BoHA fits the elementwise
ratio $E \oslash W_0$ through per-block approximations at matched
LoRA-equivalent total rank. A synthetic target where $E \oslash W_0$
has per-block rank $1$ and global rank $b^2$ separates this family
from a single global Hadamard factorization, as illustrated in
\autoref{fig:hero}(a).

Across theory, synthetic reconstruction, matched-budget single-task
evaluation, and a commonsense\,$\to$\,arithmetic continual-learning
diagnostic, the results identify spatial support as a structural axis
for $W_0$-coupled
Hadamard adaptation with measurable retention impact at matched
second-stage plasticity. Our main contributions are summarized as
follows:
\begin{itemize}[leftmargin=*, itemsep=2pt]
  \item We identify a global Hadamard parameterization shared by
    current Hadamard-product PEFT methods and introduce BoHA, a
    blockwise $W_0$-coupled Hadamard product adapter that treats
    spatial support as an explicit design axis at LoRA-equivalent
    total rank with no additional inference latency.
  \item We characterize BoHA through a per-block approximation bound
    and a gradient-locality property that separate blockwise from
    global $W_0$-coupled Hadamard parameterization.
  \item On commonsense and arithmetic benchmarks across four
    base-model scales, BoHA improves over LoRA on every
    matched-budget single-task average and remains competitive with
    the strongest Hadamard baseline.
  \item On a Llama-3.2-3B commonsense\,$\to$\,arithmetic
    continual-learning diagnostic, $W_0$-coupled Hadamard adapters
    occupy the high-retention end of the design space, with BoHA
    exceeding the $W_0$-free additive-control mean by $15.23\%$ at
    matched second-stage plasticity.
\end{itemize}

\section{Preliminaries and Design Axes}
\label{sec:design_space}

PEFT parameterizations for a frozen weight
$W_0 \in \mathbb{R}^{m \times n}$ differ along two structural axes.
The first is \emph{$W_0$-coupling}, whether the learned update is
parameterized as a multiplicative modulation of the frozen weight,
such as $\Delta W = W_0 \odot H$ for a learned matrix $H$. The second is
\emph{spatial support}, whether the learned factors span the full
matrix or act on disjoint submatrices. We separate
$W_0$-coupling from $W_0$-initialization: PiSSA~\citep{meng2024pissa}
and ABBA~\citep{singhal2026abba} can initialize from singular components
of $W_0$, but the trained update is merged additively or
through trainable factors rather than through $W_0$ itself, so the
coupling does not persist at evaluation.

Under these axes, LoRA~\citep{hu2022lora}, DoRA~\citep{liu2024dora},
and PiSSA are global, $W_0$-free additive updates. ABBA is global
and Hadamard but remains $W_0$-free because both Hadamard factors
are trainable. GraLoRA~\citep{jung2025gralora} is
blockwise but additive, and HiRA~\citep{huang2025hira} is
$W_0$-coupled but global. The proposed BoHA occupies the blockwise
$W_0$-coupled cell by placing an independent low-rank Hadamard
product inside each block of a $b{\times}b$ partition of $W_0$,
as shown in \autoref{fig:hero}(b).

The two axes play different roles in our evidence. The forward
continual-learning diagnostic compares BoHA with HiRA and with
$W_0$-free controls, including the blockwise $W_0$-free control
GraLoRA.
The theory below isolates the blockwise approximation effect. Let
$W^\star$ denote the full fine-tuning target and
$E = W^\star - W_0$ the corresponding update. The global Hadamard
parameterization, with HiRA as its $W_0$-coupled instance, fits a
single rank-$r$ factor pair to the elementwise ratio
$E \oslash W_0$ across the entire layer. When this ratio is low-rank
within local blocks but high-rank globally, the global approximation
becomes restrictive. BoHA applies the $W_0$-coupled approximation
within each block, using per-block factors at the same
LoRA-equivalent total rank. The synthetic instance in
\autoref{fig:hero}(a) illustrates this separation: a global Hadamard
parameterization matches the construction only at global rank $b^2$,
which is $b$ times BoHA's LoRA-equivalent total rank in this
synthetic example. The next section formalizes this with a blockwise
approximation bound and a gradient-locality result.

Tensor-product factorizations such as KronA~\citep{edalati2022krona}
and LoKr~\citep{yeh2024lokr} are $W_0$-free tensor-structured
updates rather than disjoint blockwise $W_0$ modulations. Related
rank-scaling and spectral analyses~\citep{kalajdzievski2023rslora,shuttleworth2025intruder}
and parameter-efficient continual fine-tuning
methods~\citep{peft_cl_survey_2025,ewc_peft_cl} are surveyed in the
related-work appendix.

\section{Method: Blockwise Hadamard Product Adaptation}
\label{sec:method}
\label{sec:theory}

For a frozen linear weight $W_0 \in \mathbb{R}^{m \times n}$, BoHA
parameterizes the trainable update as a $W_0$-coupled Hadamard
modulation applied independently over a $b{\times}b$ block partition.
At LoRA-equivalent total rank $r_{\mathrm{tot}}=b r_b$, with
per-block rank $r_b$, BoHA exactly matches LoRA's trainable-parameter
count at rank $r_{\mathrm{tot}}$ and introduces no additional
inference latency. Full derivations of the two propositions are in
Appendix~\ref{app:proofs}. Figure~\ref{fig:method_arch} locates BoHA
against LoRA, HiRA, and ABBA in the design space of
\autoref{sec:design_space}.

\begin{figure*}[t]
\centering
\includegraphics[width=0.95\linewidth]{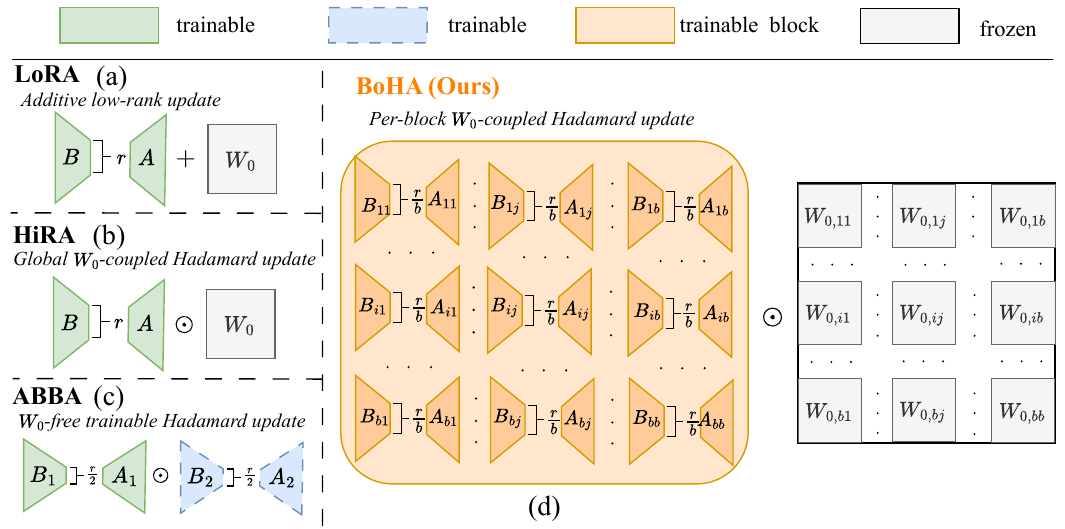}
\caption{Update parameterizations of LoRA, HiRA, ABBA, and BoHA along
the $W_0$-coupling and spatial-support design axes. BoHA partitions
$W_0$ into a $b{\times}b$ grid and learns an independent $W_0$-coupled
Hadamard modulation in each block, with $b=8$ in our main experiments.}
\label{fig:method_arch}
\end{figure*}

\textbf{Parameterization.}
HiRA~\citep{huang2025hira} represents the update as
$\Delta W_{\text{HiRA}} = W_0 \odot (B A)$ with
$A \in \mathbb{R}^{r \times n}$ and $B \in \mathbb{R}^{m \times r}$.
Let $E = W^\star - W_0$ denote a full-finetuning target update and
$\oslash$ the elementwise division operator. The HiRA
parameterization then approximates the elementwise ratio
$E \oslash W_0$ by a single global rank-$r$ product across the
entire layer. When this target is low-rank within blocks but
high-rank globally, the global factorization can become the limiting
approximation, as illustrated in \autoref{fig:hero}(a). BoHA therefore
partitions the row and column dimensions of $W_0$ into a conformal
$b{\times}b$ grid. The formulation only requires blocks
$W_{0,ij} \in \mathbb{R}^{m_i \times n_j}$ that tile $W_0$. We use
equal-size blocks in our experiments.
For each block $(i,j)$, BoHA stores factors
$A_{ij} \in \mathbb{R}^{r_b \times n_j}$,
$B_{ij} \in \mathbb{R}^{m_i \times r_b}$, assembled as
\begin{equation}
\label{eq:bhra_update}
\Delta W_{\text{BoHA}} \;=\; W_0 \odot \bigoplus_{i,j=1}^{b} B_{ij} A_{ij},
\end{equation}
where $\bigoplus$ places each $B_{ij} A_{ij}$ in its $(i,j)$ tile and
$r_{\mathrm{tot}} = b \cdot r_b$ is the LoRA-equivalent total rank.
The trainable parameter count is
$\sum_{i,j} r_b(m_i+n_j) = b r_b(m+n) = (m+n)r_{\mathrm{tot}}$,
matching LoRA at rank $r_{\mathrm{tot}}$. The sum of
per-block rank capacities is $b^2 r_b$, which enters the capacity bound
of \autoref{tab:theory_bounds} without changing the trainable-parameter
count.

This parameterization separates BoHA from the closest PEFT families. Setting
$b{=}1$ recovers HiRA exactly, and
Appendix~\ref{app:boundary_cases} reports the corresponding equivalence
check. Removing the $W_0$ factor gives a $W_0$-free blockwise additive
update, analogous to GraLoRA~\citep{jung2025gralora}. ABBA learns
both Hadamard factors~\citep{singhal2026abba}, whereas BoHA keeps the
pretrained weight frozen and trains only blockwise low-rank
modulations coupled to it.

\textbf{Blockwise approximation bound.}
Inheriting the partition above, $E$ admits the conformal partition
$E_{ij}$. Let $R_{ij} = E_{ij} \oslash W_{0,ij}$ denote the per-block
elementwise ratio and $\|M\|_{\max} = \max_{u,v} |M_{uv}|$ the
entrywise maximum-absolute norm. HiRA is the $b = 1$ special case with
$R = E \oslash W_0$. The bound is most favorable when $R$ has lower rank
within blocks than globally. The next bound replaces one global
truncation error with a sum of blockwise truncation errors.

\begin{proposition}[Blockwise approximation bound]
\label{prop:piecewise}
Assume $W_{0,ij}$ has no zero entries for every $(i,j)$, satisfied by
the dense pretrained transformer weights used in our experiments. Then
\begin{equation}
\label{eq:piecewise_bound}
\inf_{\substack{A_{ij}, B_{ij} \\ \operatorname{rank}(B_{ij}A_{ij}) \leq r_b}}
  \big\|\Delta W_{\text{BoHA}} - E\big\|_F^2
\;\leq\; \sum_{i,j=1}^{b}
  \|W_{0,ij}\|_{\max}^{2}\; \sum_{k > r_b} \sigma_k^2\!\left(R_{ij}\right),
\end{equation}
where $\sigma_k$ are singular values in decreasing order.
\end{proposition}

HiRA constrains the rank of the full $R = E \oslash W_0$ to the total
budget $r_{\mathrm{tot}}$. The bound above replaces this global
requirement with a per-block rank cap of $r_b$ at the same total
budget $r_{\mathrm{tot}} = b \cdot r_b$. \autoref{tab:theory_bounds}
lists the resulting rank upper bounds for each PEFT family. The
synthetic construction in Appendix~\ref{app:synthetic} gives the
extreme case: every block of $R$ has rank $1$ while the full $R$ has
rank $b^2$. BoHA at $r_b = 1$ reconstructs the target to machine
precision while HiRA needs global rank at least $b^2$.
The two bounds tighten under different regimes: BoHA's bound is
tight when each block of $R$ has rank at most $r_b$, while HiRA's
bound is tight when $R$ has global rank at most $r_{\mathrm{tot}}$,
including when that rank concentrates within a single block.
Empirical comparison across realistic fine-tuning targets follows
in the matched-budget single-task and continual-learning
diagnostics of Sections~\ref{sec:main_results} and~\ref{sec:cl}.

A corresponding algebraic update-rank inequality follows from the
standard Hadamard rank inequality,
$\operatorname{rank}(P \odot Q) \leq
\operatorname{rank}(P)\operatorname{rank}(Q)$
\citep{huang2025hira}. Applying it to~\eqref{eq:bhra_update}
with $C = \bigoplus_{i,j=1}^{b} B_{ij} A_{ij}$ and using rank
subadditivity over the tiled matrix gives
$\operatorname{rank}(C) \leq b^2 r_b = b \cdot r_{\mathrm{tot}}$.
Therefore
$\operatorname{rank}(\Delta W_{\text{BoHA}}) \leq b \cdot
\operatorname{rank}(W_0) \cdot r_{\mathrm{tot}}$, a factor $b$ above
HiRA at matched trainable parameters in \autoref{tab:theory_bounds}.

\begin{table}[t]
\centering
\caption{Comparison of approximation requirements and algebraic rank bounds for PEFT methods.}
\label{tab:theory_bounds}
\footnotesize
\setlength{\tabcolsep}{4.5pt}
\begin{tabular}{l@{\hspace{4pt}}lll}
\toprule
\textbf{Method} & $\operatorname{rank}(\Delta W)$ \textbf{upper bound} & \textbf{Approximation requirement} & \textbf{Dependence on $W_0$} \\
\midrule
LoRA    & $r_{\mathrm{tot}}$                        & $\operatorname{rank}(E) \leq r_{\mathrm{tot}}$                                & None \\
HiRA    & $\operatorname{rank}(W_0) \cdot r_{\mathrm{tot}}$ & $\operatorname{rank}(E \oslash W_0) \leq r_{\mathrm{tot}}$                    & Global modulation \\
GraLoRA & $b \cdot r_{\mathrm{tot}}$        & per-block $\operatorname{rank}(E_{ij}) \leq r_b$              & None \\
ABBA    & $r_1 \cdot r_2$            & $\operatorname{rank}(E) \leq r_1 r_2$                         & None (init only) \\
\rowcolor{cyan!8}
\textbf{BoHA (ours)} & $b \cdot \operatorname{rank}(W_0) \cdot r_{\mathrm{tot}}$ & \textbf{per-block} $\operatorname{rank}(E_{ij} \oslash W_{0,ij}) \leq r_b$ & Blockwise modulation \\
\bottomrule
\end{tabular}
\end{table}

\textbf{Training and inference.}
During training, the base weight $W_0$ remains frozen and only the
block factors $(A_{ij},B_{ij})$ are updated. We initialize one factor
in each block pair at zero so the initial update is zero. At
deployment, BoHA precomputes and merges the blockwise update into the
frozen weight on each block as
$W'_{ij} = W_{0,ij} \odot \bigl(\mathbf{1} + B_{ij} A_{ij}\bigr)$,
recovering a single dense matrix $W'$. As
with LoRA~\citep{hu2022lora} and HiRA~\citep{huang2025hira},
inference uses the merged weight directly and introduces no
additional inference latency.

\textbf{Blockwise gradient locality.}
The same partition also determines how gradients reach the adapter
factors. Each block sees the corresponding block of the dense layer
gradient, weighted elementwise by the frozen pretrained block.

\begin{proposition}[Blockwise gradient locality]
\label{prop:gradlocal}
Let $L$ denote the training loss. Let $g \in \mathbb{R}^{m}$ denote
the output gradient at a layer and $x \in \mathbb{R}^{n}$ its input.
Let $G = g x^\top$ and denote its $(i,j)$-block by $G_{ij}$.
Then
\begin{align}
\nabla_{A_{ij}} L &= B_{ij}^\top \bigl(W_{0,ij} \odot G_{ij}\bigr), \\
\nabla_{B_{ij}} L &= \bigl(W_{0,ij} \odot G_{ij}\bigr) A_{ij}^\top.
\end{align}
\end{proposition}

The update at block $(i,j)$ depends only on $W_{0,ij}$ and $G_{ij}$,
so a perturbation in input block $j$ affects only the $b$
blocks in column $j$. By contrast, LoRA's gradient
$\nabla_A L = B^\top G$ mixes the full outer product through full-matrix
factors. This gives a gradient-side analogue of the blockwise approximation
argument above.

\section{Experiments}
\label{sec:experiments}

We evaluate BoHA with a continual-learning diagnostic and
matched-budget single-task controls on commonsense and arithmetic
reasoning across four base-model scales. The diagnostic tests
retention under matched second-task plasticity. The single-task
controls assess matched-budget accuracy across task families.

\subsection{Datasets}
\label{sec:datasets}

\textbf{Commonsense reasoning.}
We use the eight datasets with predefined train and test splits
from LLM-Adapters~\citep{hu2023llm}, combining 170{,}420 query-answer
pairs for fine-tuning and reserving 120 entries as a validation set.
The eight datasets are BoolQ~\citep{clark2019boolq},
PIQA~\citep{bisk2020piqa}, SIQA~\citep{sap2019socialiqa},
HellaSwag~\citep{zellers2019hellaswag},
WinoGrande~\citep{sakaguchi2021winogrande}, ARC-Challenge and
ARC-Easy~\citep{clark2018think}, and OBQA~\citep{mihaylov2018can}. We
evaluate on each dataset independently and report the average across
the eight tasks to capture task-specific generalization.

\textbf{Arithmetic reasoning.}
We fine-tune Mistral-7B~\citep{mistral7b} and
Gemma-2-9B~\citep{gemma2} on a 20K-sample subset of
MetaMathQA~\citep{yu2023metamath}, and evaluate on
GSM8K~\citep{gsm8k} and MATH~\citep{math_dataset}, reporting
exact-match accuracy consistent with prior
work~\citep{singhal2026abba}.

\textbf{Code generation.}
We report pass@1 on HumanEval~\citep{chen2021humaneval} and
HumanEval+ in Appendix~\ref{app:code_gen}.

\textbf{Continual-learning diagnostic.}
We chain two task families: commonsense\,$\to$\,arithmetic
(forward) and arithmetic\,$\to$\,commonsense (reverse), training
each stage from the previous stage's merged weights. The diagnostic
reports stage-1 retention and stage-2 plasticity.

\subsection{Experimental Settings}
\label{sec:settings}
\label{sec:setup}
\label{sec:protocol}

\textbf{Baselines.}
We compare against representative PEFT methods under matched
parameter budgets: LoRA~\citep{hu2022lora}, DoRA~\citep{liu2024dora},
HiRA~\citep{huang2025hira}, GraLoRA~\citep{jung2025gralora},
PiSSA~\citep{meng2024pissa}, rsLoRA~\citep{kalajdzievski2023rslora},
and ABBA~\citep{singhal2026abba}. Full fine-tuning (FFT) is included
as a full-parameter reference. Our experiments use the
Llama-3.2-1B, Llama-3.2-3B~\citep{llama32}, Mistral-7B~\citep{mistral7b},
and Gemma-2-9B~\citep{gemma2} open-source models.

\textbf{Metrics.}
For commonsense reasoning, we use accuracy as in
HiRA~\citep{huang2025hira} and ABBA~\citep{singhal2026abba}. We parse
model completions with task-specific answer tokens, mark unmatched
responses as incorrect, and report the unweighted average over the
eight tasks. Arithmetic benchmarks follow
the GSM8K/MATH exact-match protocol after string normalization.
For the continual-learning diagnostic, we reuse per-task accuracy and
report GEM-style $R_{i,j}$, backward transfer, and post-sequence
average accuracy~\citep{lopezpaz2017gem}.

\textbf{Implementation details.}
BoHA uses total rank
$r_{\mathrm{tot}}{=}32$. We train for 2 epochs and 1 epoch on the
commonsense and arithmetic tasks, respectively. We report the mean
accuracy across multiple random seeds for each headline-table
configuration. In BoHA, we set $r_b{=}r_{\mathrm{tot}}/b$ to exactly match LoRA and
HiRA in trainable parameter count and FLOPs. We use $b{=}8$,
$r_b{=}4$ in all main comparisons. We use
$\alpha{=}r_{\mathrm{tot}}{=}32$ and initialize one factor in each
block pair at zero so that the initial update is zero, matching the
standard PEFT initialization~\citep{hu2022lora,huang2025hira}. As in
LoRA and HiRA, $\alpha$ is a method-internal constant and is not
tuned. For BoHA, we use AdamW with
learning rate $2 \times 10^{-3}$ and 100 warm-up steps. Additional
implementation details are given in Appendix~\ref{app:implementation}.

\subsection{Results on Continual Learning Diagnostic}
\label{sec:cl}

\begin{table}[t]
\centering
\caption{Forward continual-learning diagnostic on Llama-3.2-3B.}
\label{tab:cl_matrix_3b_forward}
\small
\setlength{\tabcolsep}{5pt}
\begin{tabular}{lccccc}
\toprule
\textbf{Method} & $R_{1,1}$ (\%, $\uparrow$) & $R_{2,1}$ (\%, $\uparrow$) & $R_{2,2}$ (\%, $\uparrow$) & BWT (\%, $\uparrow$) & ACC (\%, $\uparrow$) \\
\midrule
LoRA    & 81.21             & 38.55             & \underline{34.70} & $-42.66$             & 36.63             \\
DoRA    & 81.55             & 40.64             & \textbf{34.86}    & $-40.91$             & 37.75             \\
PiSSA   & 81.20             & 36.73             & 30.36             & $-44.47$             & 33.55             \\
GraLoRA & 82.63             & 53.81             & 34.29             & $-28.81$             & 44.05             \\
ABBA    & \underline{84.12} & 53.58             & 31.76             & $-30.54$             & 42.67             \\
HiRA    & 83.27             & \underline{56.86} & 31.30             & \textbf{$-26.41$}    & \underline{44.08} \\
\rowcolor{cyan!8}
BoHA    & \textbf{84.25}    & \textbf{57.66}    & 33.72             & \underline{$-26.59$} & \textbf{45.69}    \\
\bottomrule
\end{tabular}
\end{table}

\begin{wrapfigure}{r}{0.50\linewidth}
\centering
\vspace{-1em}
\includegraphics[width=\linewidth]{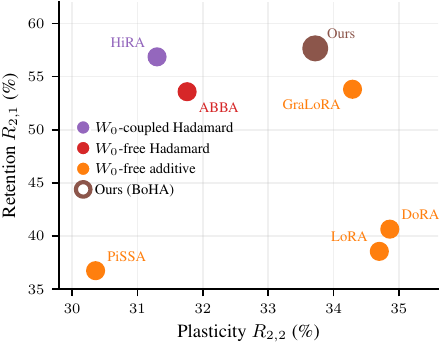}
\caption{Stability-plasticity trade-off on Llama-3.2-3B forward CL.}
\label{fig:scatter_3b}
\vspace{-1em}
\end{wrapfigure}

We chain the two task families and report the $2{\times}2$
accuracy matrix $R_{i,j}$, where $R_{i,j}$ is the stage-$i$ model's
accuracy on task $j$. The summary metrics are stage-1 retention
$R_{2,1}$, stage-2 plasticity $R_{2,2}$, backward transfer
$\mathrm{BWT} = R_{2,1} - R_{1,1}$, and post-sequence average
$\mathrm{ACC} = (R_{2,1} + R_{2,2})/2$. As shown in
Table~\ref{tab:cl_matrix_3b_forward} and
Figure~\ref{fig:scatter_3b}, the Llama-3.2-3B
commonsense\,$\to$\,arithmetic chain provides the main method-level
comparison. BoHA has the highest $R_{1,1}$,
$R_{2,1}$, and $\mathrm{ACC}$: $R_{1,1}=84.25\%$,
$R_{2,1}=57.66\%$, and $\mathrm{ACC}=45.69\%$. Relative to the $W_0$-free additive mean,
BoHA gains $15.23\%$ in retained stage-1 accuracy while changing
stage-2 accuracy by only $0.17\%$.
HiRA also retains strongly at $R_{2,1}=56.86\%$, while the strongest
$W_0$-free controls, GraLoRA and ABBA, reach $53.81\%$ and
$53.58\%$. The diagnostic places $W_0$-coupled Hadamard adapters at
the high-retention end of the design space under matched second-stage
plasticity. Reverse-direction scale checks in
Appendix~\ref{app:cross_scale} show stronger HiRA/BoHA retention at
9B and no fixed HiRA/BoHA ordering at 1B.

\subsection{Results on Commonsense Reasoning}
\label{sec:single_task_cs}
\label{sec:results}
\label{sec:main_results}

\begin{table}[t]
\centering
\caption{Results (\%) on Llama-3.2 1B and 3B across eight
commonsense reasoning datasets. The best and second-best PEFT results
within each model are indicated in \textbf{bold} and
\underline{underline}, respectively. FFT is included as a
full-parameter reference.
\textsuperscript{\textdagger} indicates results taken from
ABBA~\citep{singhal2026abba}.}
\label{tab:single_task_cs}
\small
\setlength{\tabcolsep}{2pt}
\begin{tabular}{llrccccccccc}
\toprule
\multirow{2}{*}{\textbf{Model}} & \multirow{2}{*}{\textbf{Method}} & \multirow{2}{*}{\textbf{\#Params}} & \multicolumn{9}{c}{\textbf{Accuracy ($\uparrow$)}} \\
\cmidrule(lr){4-12}
 & & & \textbf{OBQA} & \textbf{ARC-c} & \textbf{ARC-e} & \textbf{Wino} & \textbf{HellaS} & \textbf{PIQA} & \textbf{SIQA} & \textbf{BoolQ} & \textbf{Avg.} \\
\midrule
\multirow{9}{*}{\rotatebox[origin=c]{90}{Llama-3.2-1B}}
 & FFT                                              & 1.24B  & 74.00 & 62.05 & 78.63 & 74.79 & 79.63 & 80.62 & 75.37 & 63.77 & 73.61 \\
 & LoRA                       & 22.54M & 71.40 & 60.49 & 76.56 & 72.38 & 77.41 & 78.78 & 73.39 & 65.11 & 71.94 \\
 & rsLoRA\textsuperscript{\textdagger}               & 22.54M & 71.11 & 59.85 & 74.90 & 73.85 & 75.34 & 78.32 & 73.47 & \underline{65.44} & 71.54 \\
 & DoRA                       & 22.92M & 72.20 & 61.09 & 76.77 & 73.01 & 77.48 & 78.45 & 73.34 & 64.83 & 72.15 \\
 & PiSSA                      & 22.54M & 72.20 & 61.43 & 77.40 & \underline{74.03} & 76.82 & 79.05 & \textbf{75.54} & 65.20 & 72.71 \\
 & GraLoRA                    & 22.54M & 71.40 & 60.92 & 78.54 & 73.95 & 77.81 & 79.87 & 73.95 & 65.02 & 72.68 \\
 & ABBA            & 22.54M & \underline{72.93} & \underline{62.29} & \underline{79.00} & \textbf{74.74} & \textbf{80.99} & \underline{80.00} & \underline{74.75} & 65.37 & \textbf{73.76} \\
 & HiRA                       & 22.54M & 70.20 & 58.70 & 77.61 & 71.27 & 75.74 & 79.22 & 73.39 & 65.26 & 71.42 \\
\rowcolor{cyan!8}
\cellcolor{white} & BoHA                       & 22.54M & \textbf{73.33} & \textbf{62.34} & \textbf{79.89} & 73.59 & \underline{79.82} & \textbf{80.59} & 73.83 & \textbf{65.55} & \underline{73.62} \\
\midrule
\multirow{9}{*}{\rotatebox[origin=c]{90}{Llama-3.2-3B}}
 & FFT                                              & 3.21B  & 85.00 & 78.81 & 90.00 & 86.55 & 93.14 & 87.25 & 81.49 & 73.58 & 84.48 \\
 & LoRA                       & 48.63M & 79.60 & 74.32 & 87.54 & 83.50 & 89.63 & 85.20 & 79.84 & 70.00 & 81.20 \\
 & rsLoRA\textsuperscript{\textdagger}               & 48.63M & 81.72 & 74.18 & 86.71 & 82.02 & 90.45 & 85.05 & 78.92 & 69.81 & 81.11 \\
 & DoRA                       & 49.40M & 79.80 & 74.15 & 88.09 & 84.14 & 90.43 & 85.47 & 79.84 & 70.24 & 81.52 \\
 & PiSSA                      & 48.63M & 80.80 & 73.52 & 86.77 & 83.24 & 88.05 & 84.71 & 79.43 & 72.63 & 81.17 \\
 & GraLoRA                    & 48.63M & 83.73 & 76.00 & 88.90 & 84.98 & 92.38 & 86.53 & \underline{81.03} & 72.28 & 83.23 \\
 & ABBA            & 48.63M & \underline{84.27} & \underline{77.87} & \textbf{89.66} & \textbf{86.53} & \textbf{93.49} & \textbf{87.20} & 81.00 & \underline{73.00} & \underline{84.13} \\
 & HiRA                       & 48.63M & 83.32 & 77.61 & 89.53 & 85.71 & 92.94 & 86.57 & 80.81 & 72.83 & 83.67 \\
\rowcolor{cyan!8}
\cellcolor{white} & BoHA                       & 48.63M & \textbf{85.73} & \textbf{78.78} & \underline{89.62} & \underline{86.40} & \underline{93.16} & \underline{86.69} & \textbf{81.35} & \textbf{74.19} & \textbf{84.49} \\
\bottomrule
\end{tabular}
\end{table}

Table~\ref{tab:single_task_cs} reports matched-budget commonsense
results. BoHA improves over the corresponding LoRA row by $+1.68$
percentage points on Llama-3.2-1B and $+3.29$ percentage points on
Llama-3.2-3B. It matches the FFT average within $0.01$ percentage points at both
scales while using under $2\%$ of the trainable parameter count. Compared
with ABBA, BoHA is below ABBA by $0.14$ percentage points at 1B and
above ABBA by $0.36$ percentage points at 3B, indicating competitive
matched-budget single-task capacity within the Hadamard family.

\subsection{Results on Arithmetic Reasoning}
\label{sec:single_task_arith}

\begin{table}[t]
\centering
\caption{Results (\%) on Mistral-7B and Gemma-2-9B
across arithmetic reasoning benchmarks.
\textsuperscript{\textdagger} indicates results taken from ABBA~\citep{singhal2026abba}.}
\label{tab:single_task_arith}
\small
\setlength{\tabcolsep}{3pt}
\begin{tabular}{lrccrcc}
\toprule
\multirow{2}{*}{\textbf{Method}}
 & \multicolumn{3}{c}{\textbf{Mistral-7B}} & \multicolumn{3}{c}{\textbf{Gemma-2-9B}} \\
\cmidrule(lr){2-4}\cmidrule(lr){5-7}
 & \textbf{\#Params} & \textbf{GSM8K ($\uparrow$)} & \textbf{MATH ($\uparrow$)} & \textbf{\#Params} & \textbf{GSM8K ($\uparrow$)} & \textbf{MATH ($\uparrow$)} \\
\midrule
FFT\textsuperscript{\textdagger}    &   7.24B & 63.87          & 17.65            &   9.24B & 79.23            & 38.02            \\
LoRA         &  83.89M & 63.15          & 15.28            & 108.04M & 76.40            & 34.68            \\
rsLoRA\textsuperscript{\textdagger} &  83.88M & 62.15          & 16.24            & 108.04M & 76.84            & 36.88            \\
DoRA    &  85.26M & 63.76          & 17.90            & 109.89M & 78.22            & 37.76            \\
PiSSA\textsuperscript{\textdagger}   &  83.88M & 62.43          & 16.52            & 108.04M & 77.12            & 37.04            \\
ABBA    &  83.89M & \textbf{64.37} & \underline{17.96} & 108.04M & 78.39             & \underline{38.50} \\
HiRA    &  83.89M & 62.52          & 17.31            & 108.04M & \underline{78.44} & \textbf{38.93}    \\
\rowcolor{cyan!8}
BoHA    &  83.89M & \underline{64.11} & \textbf{17.97} & 108.04M & \textbf{79.15}    & 38.44            \\
\bottomrule
\end{tabular}
\end{table}

Table~\ref{tab:single_task_arith} reports matched-budget arithmetic
results. BoHA reaches $17.97\%$ on Mistral-7B MATH versus ABBA's
$17.96\%$, and is highest on Gemma-2-9B GSM8K at $79.15\%$. ABBA is
highest on Mistral-7B GSM8K and HiRA is highest on Gemma-2-9B MATH. Across these model
and benchmark pairs, BoHA remains close to the strongest Hadamard
baseline while improving over LoRA in both base-model settings.

\subsection{Training Cost and Memory}
\label{sec:efficiency}

\begin{table}[t]
\centering
\caption{Peak GPU memory and training wall-clock at matched
trainable-parameter budget on Llama-3.2-3B commonsense reasoning.}
\label{tab:gram_runtime}
\footnotesize
\setlength{\tabcolsep}{8pt}
\begin{tabular}{lcc}
\toprule
\textbf{Configuration} & \textbf{Peak GPU memory (GB)} & \textbf{Training wall-clock} \\
\midrule
LoRA ($r{=}32$)              & $21.16$ & $3$ h $31$ m $45$ s \\
HiRA ($r{=}32$)              & $21.16$ & $3$ h $52$ m $34$ s \\
\rowcolor{cyan!8}
BoHA ($b{=}8$, $r_b{=}4$)    & $21.16$ & $3$ h $52$ m $41$ s \\
\bottomrule
\end{tabular}
\end{table}

Table~\ref{tab:gram_runtime} reports wall-clock and peak memory for
matched-budget Llama-3.2-3B commonsense runs using the same batch,
sequence length, and two-epoch schedule. Peak memory is identical
across LoRA, HiRA, and BoHA in this measurement, consistent with
frozen-model activations dominating adapter state at this scale. BoHA
matches HiRA's training time in this implementation and adds $20$ minutes
$56$ seconds over LoRA. BoHA uses one dense weight matrix per adapted
projection and introduces no additional inference latency.

\section{Analysis}
\label{sec:analysis}

The experiments above show that BoHA remains competitive in
matched-budget single-task tables. The forward continual-learning
diagnostic places $W_0$-coupled Hadamard adapters at the
high-retention end of the design space. This section
analyzes four facets of BoHA's parameterization: the scope of the
retention signal, block-count and rank-budget robustness,
trained-update spectra, and component placement in the transformer.

\subsection{Scope of the Retention Diagnostic}
\label{sec:cl_mechanism}
\label{sec:mechanistic}

The forward continual-learning diagnostic directly compares
method-level behavior. On the Llama-3.2-3B
commonsense$\rightarrow$arithmetic sequence, BoHA achieves
$R_{2,1}=57.66\%$ and $R_{2,2}=33.72\%$ in
Table~\ref{tab:cl_matrix_3b_forward}, a $15.23\%$ gain in retained
stage-1 accuracy over the $W_0$-free additive mean while changing
stage-2 accuracy by only $0.17\%$. The same
table separates $W_0$-coupling and spatial support. HiRA, the $b{=}1$
case of BoHA, also retains strongly. GraLoRA is the blockwise
$W_0$-free control and ABBA is the global $W_0$-free Hadamard
baseline. The diagnostic supports the $W_0$-coupled Hadamard
component shared by BoHA and HiRA, while the following subsections
examine the blockwise-localization component.

\subsection{Block Count and Rank-Budget Robustness}
\label{sec:rb_choice}

\begin{wrapfigure}{r}{0.50\linewidth}
\centering
\vspace{-0.8em}
\includegraphics[width=\linewidth]{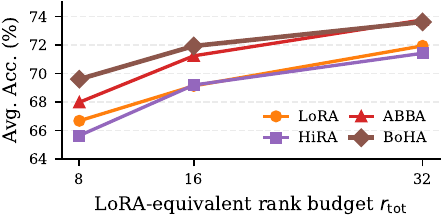}
\caption{Rank-budget diagnostic on Llama-3.2-1B commonsense reasoning.}
\label{fig:rank_budget}
\vspace{-0.8em}
\end{wrapfigure}

Figure~\ref{fig:rank_budget} reports a rank-budget diagnostic on
Llama-3.2-1B, with full per-task results in
Table~\ref{tab:appendix_rank_budget}. BoHA is above ABBA by $1.62$
percentage points at $r_{\mathrm{tot}}{=}8$ and $0.69$ percentage
points at $r_{\mathrm{tot}}{=}16$, and ABBA is above BoHA by $0.14$
percentage points at $r_{\mathrm{tot}}{=}32$, supporting low-budget
competitiveness within the Hadamard family. At fixed total budget
$r_{\mathrm{tot}} = b\,r_b$, smaller $r_b$ tightens BoHA's per-block
approximation requirement.
Table~\ref{tab:block_count_sweep_3b} reports the Llama-3.2-3B
block-count sweep at $r_{\mathrm{tot}}{=}32$, with BoHA's average
accuracy rising from $83.50\%$ to $83.98\%$ and $84.49\%$ for
$b{=}2,4,8$. ABBA reaches $84.13\%$ in
Table~\ref{tab:single_task_cs}, above $b\in\{2,4\}$ but below
$b{=}8$. We therefore use $b{=}8$, $r_b{=}4$ in all main results as
the strongest tested setting under the shared training protocol.

\subsection{Singular Value Structure of Trained Updates}
\label{sec:spectral}

\begin{figure}[t]
\centering
\includegraphics[width=\linewidth,height=0.115\textheight,keepaspectratio]{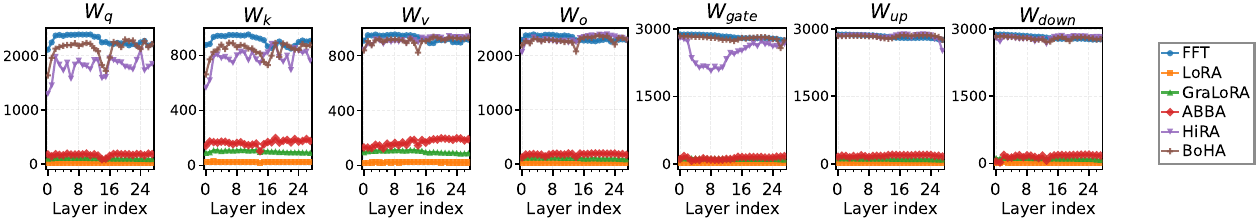}
\caption{Entropy-based effective rank~\citep{roy2007effective}
across layers for FFT, LoRA, GraLoRA, ABBA, HiRA, and BoHA at
Llama-3.2-3B.}
\label{fig:effective_rank}
\end{figure}

The bounds in Table~\ref{tab:theory_bounds} describe feasible update
families. We measure realized spectra of trained updates at
Llama-3.2-3B across the seven projections against the FFT proxy
$\bar{E} = W_{\text{FFT}} - W_0$~\citep{huang2025hira}.
Figure~\ref{fig:effective_rank} reports the entropy-based effective
rank~\citep{roy2007effective} per layer. Effective-rank ordering
follows the $W_0$-coupled versus $W_0$-free grouping: the
$W_0$-coupled methods cluster near FFT, with BoHA at $2083$, HiRA at
$1982$, and FFT at $2130$, while the decoupled Hadamard baseline
ABBA remains bounded by $r_1 r_2$ at $\sim 164$. BoHA exceeds HiRA
in mean effective rank by $+19\%$ on $W_q$, $+7\%$ on $W_k$, and
$+13\%$ on $W_{\mathrm{gate}}$, and is tied within $\pm 1\%$
elsewhere.

\begin{figure}[t]
\centering
\includegraphics[width=\linewidth,height=0.115\textheight,keepaspectratio]{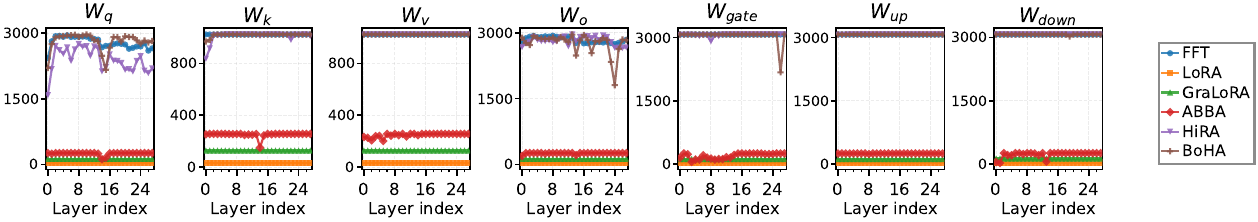}
\caption{Count of singular values exceeding $1\%$ of the
layer-wise maximum for FFT, LoRA, GraLoRA, ABBA, HiRA, and BoHA at
Llama-3.2-3B.}
\label{fig:sv_counts}
\end{figure}

Figure~\ref{fig:sv_counts} reports the count of singular values
above $1\%$ of $\sigma_{\max}$ per layer. LoRA and GraLoRA reach
their algebraic ceilings of $32$ and $128$. For HiRA and BoHA, the
algebraic bounds in Table~\ref{tab:theory_bounds} are far above the
realized $\sim 2000$ effective rank, indicating that trained updates
concentrate well below the feasible ceiling. The corresponding
Frobenius-energy profile is in Appendix~\ref{app:sv_extra}.

\subsection{Placement of BoHA in Transformers}
\label{sec:placement}

\begin{wrapfigure}{r}{0.5\linewidth}
\centering
\vspace{-1.5em}
\includegraphics[width=\linewidth]{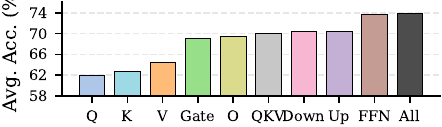}
\caption{BoHA component-placement diagnostic on Llama-3.2-1B
commonsense reasoning at matched total rank.}
\label{fig:placement}
\vspace{-1.0em}
\end{wrapfigure}

Figure~\ref{fig:placement} and Table~\ref{tab:placement_full} report
BoHA applied to individual transformer components on Llama-3.2-1B at
matched total rank. Among grouped variants, the feed-forward triple
(Up, Down, Gate) reaches $73.68\%$, which is $99.85\%$ of the
All-component average of $73.79\%$, while the joint attention input
QKV reaches $69.96\%$. The strongest single projections are Up at
$70.36\%$ and Down at $70.32\%$, and the weakest are the attention
scoring projections Q at $61.95\%$ and K at $62.66\%$. This ordering
is consistent with the placement trend reported by
HiRA~\citep{huang2025hira} and ABBA~\citep{singhal2026abba}, where
attention scoring weights contribute the least and
representation-transforming weights contribute the most. We use the
All-component placement throughout the main-table comparisons of
Sections~\ref{sec:cl} and~\ref{sec:main_results}.

\begin{table}[H]
\centering
\caption{Performance of the Llama-3.2-1B model with BoHA integrated
into various components.}
\label{tab:placement_full}
\small
\setlength{\tabcolsep}{2pt}
\begin{tabular*}{\textwidth}{@{\extracolsep{\fill}}lccccccccc@{}}
\toprule
\textbf{Component} & \textbf{OBQA} & \textbf{ARC-c} & \textbf{ARC-e} & \textbf{Wino} & \textbf{HellaS} & \textbf{PIQA} & \textbf{SIQA} & \textbf{BoolQ} & \textbf{Avg.} ($\uparrow$) \\
\midrule
All  & 73.04 & 62.66 & 79.47 & 74.02 & 80.35 & 80.47 & 74.41 & 65.91 & 73.79 \\
FFN  & 73.10 & 62.37 & 79.48 & 73.90 & 79.60 & 81.04 & 74.16 & 65.79 & 73.68 \\
Up   & 67.80 & 63.29 & 67.57 & 73.37 & 71.85 & 75.66 & 75.17 & 68.21 & 70.36 \\
Down & 68.80 & 59.00 & 76.70 & 69.97 & 73.44 & 77.48 & 72.68 & 64.50 & 70.32 \\
QKV  & 66.68 & 57.49 & 75.64 & 70.73 & 75.62 & 78.01 & 73.19 & 62.31 & 69.96 \\
O    & 66.52 & 57.13 & 75.29 & 68.92 & 74.27 & 78.16 & 72.36 & 63.60 & 69.53 \\
Gate & 65.73 & 57.17 & 76.25 & 68.48 & 71.32 & 77.37 & 71.60 & 64.18 & 69.01 \\
V    & 57.95 & 50.80 & 69.98 & 64.39 & 67.85 & 76.43 & 67.90 & 60.96 & 64.53 \\
K    & 58.50 & 48.98 & 69.70 & 63.63 & 63.21 & 74.08 & 65.90 & 57.32 & 62.66 \\
Q    & 57.20 & 48.55 & 67.06 & 65.07 & 58.54 & 73.48 & 64.69 & 60.99 & 61.95 \\
\bottomrule
\end{tabular*}
\end{table}

\section{Conclusions}
\label{sec:discussion}

We presented BoHA, a blockwise $W_0$-coupled Hadamard product adapter
for parameter-efficient fine-tuning that partitions the frozen weight
into a $b{\times}b$ grid, learns an independent low-rank factorization
in each block, and admits a per-block approximation bound and a
gradient-locality property.
BoHA remains competitive in matched-budget single-task experiments,
and the forward matched-plasticity retention diagnostic places
$W_0$-coupled Hadamard adapters at the high-retention end of the
design space, with BoHA retaining more stage-1 accuracy than the
$W_0$-free additive mean with little change in stage-2 plasticity. These results
support blockwise $W_0$-coupled Hadamard adaptation as a competitive
PEFT design choice when retention under sequential adaptation is part
of the objective.

\bibliographystyle{neurips2026/neurips_2026}
\bibliography{neurips2026/aliases,neurips2026/neurips_2026}

\newpage
\appendix
\section*{Appendix}
\section{Extended Related Work}
\label{app:related_work}

\paragraph{Tensor-product factorisations: KronA and LoKr.}
KronA~\citep{edalati2022krona} parameterises the update as a
Kronecker product $\Delta W = A \otimes B$ with
$A \in \mathbb{R}^{m_a \times n_a}$,
$B \in \mathbb{R}^{m_b \times n_b}$, and $m = m_a m_b$, $n = n_a n_b$.
LoKr~\citep{yeh2024lokr} also low-rank-factorises each
Kronecker factor, giving
$\Delta W = (B_1 A_1) \otimes (B_2 A_2)$. We separate this family
from BoHA on two grounds. First, both Kronecker variants are
$W_0$-free: the trained $\Delta W$ is added to $W_0$, not coupled to
it, so they sit in the same $W_0$-free row of the design grid as
LoRA and ABBA. Second, the Kronecker product imposes a rigid
index-tensorised structure: the $((i_a, i_b), (j_a, j_b))$-entry of
$A \otimes B$ equals $A_{i_a j_a}\, B_{i_b j_b}$, so every entry is
the product of one factor coordinate in $A$ and one coordinate in
$B$. BoHA's per-block Hadamard form
$\Delta W_{ij} = W_{0,ij} \odot (B_{ij} A_{ij})$ does not share this
constraint. Blocks are independent, each coupling is element-wise
against the corresponding $W_{0,ij}$, and the spatial support is a
disjoint partition rather than a tensor product. As a consequence,
KronA and LoKr capacity scales with the chosen Kronecker
factorisation $(m_a, n_a, m_b, n_b)$ and is not directly
commensurate with our $(b, r_b)$ budget at matched parameter count.
KronA and LoKr are therefore placed in the design grid of
\autoref{fig:hero}(b) rather than in
Tables~\ref{tab:single_task_cs} and~\ref{tab:single_task_arith}.

\paragraph{LoRA-family budget allocation and rank scaling.}
AdaLoRA~\citep{zhang2023adalora} reallocates low-rank budget across
weight matrices by importance, and
rsLoRA~\citep{kalajdzievski2023rslora} introduces $\alpha / \sqrt{r}$
scaling. BoHA applies the same scaling convention with per-block rank
$r_b$, while keeping the rank budget uniform across modules to
isolate the spatial-support axis from per-module rank reallocation.

\paragraph{Approximation bounds and spectral analyses.}
HiRA~\citep{huang2025hira} bounds its global approximation error by
the $(r+1)$-th singular value of $E \oslash W_0$, of which
Proposition~\ref{prop:piecewise} is a direct corollary.
Intruder-dimension analysis~\citep{shuttleworth2025intruder}
contrasts LoRA with full fine-tuning but does not compare PEFT
families.

\paragraph{Continual fine-tuning.}
Parameter-efficient continual fine-tuning surveys~\citep{peft_cl_survey_2025}
and regulariser-based methods~\citep{ewc_peft_cl} address forgetting
through auxiliary mechanisms rather than through the update
parameterisation.

\section{Detailed Proofs of Propositions \ref{prop:piecewise} and \ref{prop:gradlocal}}
\label{app:proofs}

We re-state and prove the two main-body propositions in full. We
write $\|M\|_{\max} = \max_{u,v}|M_{uv}|$ for the entrywise
maximum-absolute norm and $\sigma_k(M)$ for the $k$-th singular
value of $M$ in decreasing order.

\paragraph{Hadamard-Frobenius bound.}
For matrices $A, B \in \mathbb{R}^{p \times q}$ of the same shape,
\begin{equation}
\label{eq:had_frob_bound}
\|A\odot B\|_F^2 \;\leq\; \|A\|_{\max}^2 \,\|B\|_F^2 .
\end{equation}
The bound follows from
$\|A\odot B\|_F^2 = \sum_{u,v} A_{uv}^2 B_{uv}^2
\leq \max_{u,v} A_{uv}^2 \cdot \sum_{u,v} B_{uv}^2
= \|A\|_{\max}^2 \,\|B\|_F^2$.

\paragraph{Proof of Proposition~\ref{prop:piecewise}.}
The placement operator $\bigoplus$ in~\eqref{eq:bhra_update} sends
each $B_{ij}A_{ij}$ to its $(i,j)$-tile. Since $W_0$ and $E$ admit
the conformal partition with blocks $W_{0,ij}$ and $E_{ij}$, the
Frobenius norm decomposes as
\begin{equation}
\|\Delta W_{\text{BoHA}} - E\|_F^2
\;=\; \sum_{i,j=1}^{b}\bigl\|W_{0,ij}\odot(B_{ij}A_{ij}) - E_{ij}\bigr\|_F^2 .
\end{equation}
Under the assumption that $W_{0,ij}$ has no zero entries,
$R_{ij} = E_{ij}\oslash W_{0,ij}$ is well-defined and
$E_{ij} = W_{0,ij}\odot R_{ij}$. We verified empirically that no
entries are exactly zero across the
$q/k/v/o/\mathrm{gate}/\mathrm{up}/\mathrm{down}$ projections of the
four pretrained checkpoints used in our experiments (Llama-3.2 1B/3B,
Mistral-7B, Gemma-2-9B). The $(i,j)$-th term therefore
equals $\|W_{0,ij}\odot(B_{ij}A_{ij} - R_{ij})\|_F^2$, and
\eqref{eq:had_frob_bound} gives
\begin{equation}
\bigl\|W_{0,ij}\odot(B_{ij}A_{ij} - R_{ij})\bigr\|_F^2
\;\leq\; \|W_{0,ij}\|_{\max}^{2}\,
\bigl\|B_{ij}A_{ij} - R_{ij}\bigr\|_F^2 .
\end{equation}
The factor pairs $(B_{ij}, A_{ij})$ are unconstrained across
blocks, so the per-block infima decouple. By the
Eckart-Young-Mirsky theorem~\citep{eckart1936approximation,mirsky1960symmetric},
the optimal rank-$r_b$ Frobenius approximation to $R_{ij}$ is its
truncated SVD, with residual
\begin{equation}
\inf_{\substack{B_{ij}, A_{ij}\\\operatorname{rank}(B_{ij}A_{ij})\leq r_b}}
\bigl\|B_{ij}A_{ij} - R_{ij}\bigr\|_F^2
\;=\; \sum_{k>r_b}\sigma_k^2(R_{ij}) .
\end{equation}
Summing over $(i,j)$ yields~\eqref{eq:piecewise_bound}. \hfill$\square$

\paragraph{Proof of Proposition~\ref{prop:gradlocal}.}
Partition the output gradient $g = [g_1^\top,\dots,g_b^\top]^\top$
with $g_i\in\mathbb{R}^{m_i}$ and the input
$x = [x_1^\top,\dots,x_b^\top]^\top$ with $x_j\in\mathbb{R}^{n_j}$
conformally with the row and column splits of $W_0$. Then
$G = g x^\top$ has $(i,j)$-block $G_{ij} = g_i x_j^\top$.
By~\eqref{eq:bhra_update}, $\Delta W$ on the $(i,j)$-tile equals
$W_{0,ij}\odot(B_{ij}A_{ij})$, and depends only on
$(B_{ij}, A_{ij})$. Writing $z = (W_0+\Delta W)x$ and
$g = \nabla_z L$, the Frobenius differential of $L$ at fixed
$(B_{kl}, A_{kl})$ for $(k,l) \neq (i,j)$ is
\begin{equation}
dL \;=\; \bigl\langle G_{ij},\;
W_{0,ij}\odot\bigl(dB_{ij}\,A_{ij} + B_{ij}\,dA_{ij}\bigr)\bigr\rangle_F
\;=\; \bigl\langle W_{0,ij}\odot G_{ij},\;
dB_{ij}\,A_{ij} + B_{ij}\,dA_{ij}\bigr\rangle_F ,
\end{equation}
where the second equality uses the self-adjointness of the
Hadamard product under the Frobenius inner product. Matching
against $dL = \langle \nabla_{B_{ij}}L,\, dB_{ij}\rangle_F
+ \langle \nabla_{A_{ij}}L,\, dA_{ij}\rangle_F$ and applying the
matrix-product adjoint identities
$\langle X, Y A^\top\rangle_F = \langle X A, Y\rangle_F$ and
$\langle X, B^\top Y\rangle_F = \langle B X, Y\rangle_F$ yields
\begin{equation}
\nabla_{A_{ij}} L \;=\; B_{ij}^\top\bigl(W_{0,ij}\odot G_{ij}\bigr),
\qquad
\nabla_{B_{ij}} L \;=\; \bigl(W_{0,ij}\odot G_{ij}\bigr)\,A_{ij}^\top .
\end{equation}
\hfill$\square$

\section{Synthetic Reconstruction Sweep}
\label{app:synthetic}

We construct a target update on which BoHA reconstructs exactly at
$r_b{=}1$ while HiRA's global approximation matches the construction
only at $r \geq b^2$. Fix $m = n = 64$ and $b = 8$. All Gaussian draws below
use a fixed random seed for reproducibility. Let
$W_0 \in \mathbb{R}^{64 \times 64}$ have i.i.d.\ $\mathcal{N}(0,1)$
entries. For every block $(i,j)$, define a rank-$1$ matrix
$R_{ij} = u_{ij} v_{ij}^\top$ with $u_{ij}, v_{ij}$ sampled from
$\mathcal{N}(0,I_8)$, and set $E_{ij} = W_{0,ij} \odot R_{ij}$. Then
$\operatorname{rank}(R_{ij}) = 1$ for every block whereas
$\operatorname{rank}(R) = 64$ globally. BoHA at $r_b = 1$ attains
$\|\Delta W - E\|_F = 2.38 \times 10^{-14}$. HiRA at the matched
per-block budget $r = 8$ retains error $42.98$, and
Table~\ref{tab:synthetic} shows that it reaches numerical zero
only at $r = b^2 = 64$.

\begin{table}[H]
\centering
\caption{Reconstruction error on the blockwise rank-$1$ synthetic
target ($m = n = 64$, $b = 8$, $\|E\|_F = 61.37$). BoHA at
total rank capacity $64$ attains numerical zero. HiRA needs $r = 64$
to match.}
\label{tab:synthetic}
\footnotesize
\setlength{\tabcolsep}{6pt}
\begin{tabular}{lrr}
\toprule
Method (config) & Global rank & $\|\Delta W - E\|_F$ \\
\midrule
HiRA ($r=8$)                        & 8  & $42.98$ \\
HiRA ($r=16$)                       & 16 & $28.55$ \\
HiRA ($r=32$)                       & 32 & $11.26$ \\
HiRA ($r=48$)                       & 48 & $2.06$ \\
HiRA ($r=60$)                       & 60 & $0.08$ \\
HiRA ($r=64$)                       & 64 & $1.04 \times 10^{-13}$ \\
\rowcolor{cyan!8}
\textbf{BoHA} ($b=8, r_b=1$)        & 64 & $\mathbf{2.38 \times 10^{-14}}$ \\
\bottomrule
\end{tabular}
\end{table}

\section{Implementation Details}
\label{app:implementation}

We implement all methods in PyTorch~\citep{paszke2019pytorch} with
HuggingFace Transformers~\citep{wolf2020transformers}. Experiments
run on NVIDIA RTX 5090 32~GB and A100 80~GB GPUs. Base models are
loaded in \texttt{torch.bfloat16} to reduce memory consumption.
Every configuration is trained with the AdamW
optimizer~\citep{loshchilov2017decoupled} with $\beta_1 = 0.9$,
$\beta_2 = 0.999$ and weight decay $0$. We train each headline-table
configuration with multiple random seeds and report the arithmetic
mean across all completed seed runs. Fixed diagnostic runs are
reported under their stated protocol.

We configure BoHA on Llama-3.2 1B, Llama-3.2 3B, Mistral-7B, and
Gemma-2 9B using the hyperparameters in Table~\ref{tab:hyper_it}.
For baseline comparisons, we replicate the experimental setups from the original
LoRA~\citep{hu2022lora}, DoRA~\citep{liu2024dora},
PiSSA~\citep{meng2024pissa}, GraLoRA~\citep{jung2025gralora},
rsLoRA~\citep{kalajdzievski2023rslora}, HiRA~\citep{huang2025hira},
and ABBA~\citep{singhal2026abba} papers to ensure fair and
consistent evaluation.

\begin{table}[H]
\centering
\caption{Hyperparameter settings of BoHA for training Llama-3.2 1B
and 3B on \textsc{Commonsense170K}, and Mistral-7B and Gemma-2 9B on
MetaMathQA.}
\label{tab:hyper_it}
\small
\begin{tabular}{lcc}
\toprule
 & \textbf{Llama-3.2 1B / 3B} & \textbf{Mistral-7B / Gemma-2 9B} \\
\midrule
Optimizer        & AdamW                                      & AdamW \\
Batch size       & $6$                                        & $1$ \\
Max.\ Seq.\ Len  & $256$                                      & $512$ \\
Grad Acc.\ Steps & $24$                                       & $32$ \\
Epochs           & $2$                                        & $1$ \\
Dropout          & $0.05$                                     & $0$ \\
Learning Rate    & $2\times 10^{-3}$                          & $2\times 10^{-3}$ \\
Target Modules   & \multicolumn{2}{c}{\texttt{q\_proj}, \texttt{k\_proj}, \texttt{v\_proj}, \texttt{o\_proj}, \texttt{gate\_proj}, \texttt{up\_proj}, \texttt{down\_proj}} \\
LR Scheduler     & Linear                                     & Cosine \\
Warmup Ratio     & $0.02$                                     & $0.02$ \\
\bottomrule
\end{tabular}
\end{table}

For the 2-task continual-learning diagnostic we report the
$2{\times}2$ accuracy matrix $R_{i,j}$, where $R_{i,j}$ is the
accuracy of the stage-$i$ model on task
$T_j$~\citep{lopezpaz2017gem}. With $T_1$ and $T_2$ the two stages,
$R_{2,1}$, the \emph{retention} term, is the stage-2 model's accuracy on the
stage-1 task and is our primary diagnostic. $R_{2,2}$, the
\emph{plasticity} term, is its accuracy on the stage-2 task. From the
matrix we derive the standard GEM retention summaries:
\begin{align*}
  \mathrm{BWT} &= R_{2,1} - R_{1,1}
    \quad \text{(backward transfer)}, \\
  F           &= \max(0,\, R_{1,1} - R_{2,1})
              \;=\; \max(0,\, -\mathrm{BWT})
    \quad \text{(forgetting)}, \\
  \mathrm{ACC} &= (R_{2,1} + R_{2,2})/2
    \quad \text{(post-sequence average)}.
\end{align*}
$F$ is reported as a non-negative scalar: positive BWT corresponds
to backward improvement, negative BWT to forgetting. The
pre-registered $\pm 1\%$ plasticity band applies to the
family-level stage-2 gap
$\bar{R}_{2,2}^{(W_0\text{-coupled})} - \bar{R}_{2,2}^{(W_0\text{-free})}$,
averaged within each $W_0$-coupling family.

\section{Rank-Budget Diagnostic}
\label{app:rank_budget}

Table~\ref{tab:appendix_rank_budget} gives the per-task results for
the rank-budget diagnostic summarised in
Figure~\ref{fig:rank_budget}. The diagnostic varies the
LoRA-equivalent rank budget while keeping the same commonsense
training and evaluation protocol.

\begin{table}[H]
\centering
\caption{Rank-budget diagnostic on Llama-3.2-1B commonsense reasoning.
Results are reported as accuracy (\%) over the eight sub-tasks and their
average.}
\label{tab:appendix_rank_budget}
\small
\setlength{\tabcolsep}{2.5pt}
\begin{tabular}{llccccccccc}
\toprule
\textbf{$r_{\mathrm{tot}}$} & \textbf{Method}
& \textbf{OBQA} & \textbf{ARC-c} & \textbf{ARC-e}
& \textbf{Wino} & \textbf{HellaS} & \textbf{PIQA}
& \textbf{SIQA} & \textbf{BoolQ} & \textbf{Avg.} \\
\midrule
\multirow{4}{*}{8}
 & LoRA & 62.00 & 52.73 & 73.23 & 68.03 & 66.48 & 76.12 & 71.70 & 63.12 & 66.68 \\
 & HiRA & 59.60 & 52.90 & 72.52 & 65.59 & 66.06 & 75.08 & 70.06 & 63.33 & 65.64 \\
 & ABBA & 64.40 & 55.20 & 73.23 & 68.51 & 70.12 & 75.52 & 73.49 & 63.33 & 67.98 \\
\rowcolor{cyan!8}
\cellcolor{white} & BoHA & 67.40 & 56.91 & 76.30 & 68.98 & 73.10 & 77.15 & 72.52 & 64.40 & \textbf{69.60} \\
\midrule
\multirow{4}{*}{16}
 & LoRA & 65.40 & 57.25 & 74.71 & 69.93 & 72.80 & 76.55 & 72.62 & 63.79 & 69.13 \\
 & HiRA & 66.80 & 56.40 & 75.80 & 68.67 & 71.78 & 77.64 & 71.75 & 64.71 & 69.19 \\
 & ABBA & 67.80 & 59.22 & 76.64 & 71.82 & 77.44 & 78.94 & 73.29 & 64.77 & 71.24 \\
\rowcolor{cyan!8}
\cellcolor{white} & BoHA & 72.00 & 59.30 & 78.91 & 70.88 & 77.13 & 78.45 & 73.85 & 64.92 & \textbf{71.93} \\
\midrule
\multirow{4}{*}{32}
 & LoRA & 71.40 & 60.49 & 76.56 & 72.38 & 77.41 & 78.78 & 73.39 & 65.11 & 71.94 \\
 & HiRA & 70.20 & 58.70 & 77.61 & 71.27 & 75.74 & 79.22 & 73.39 & 65.26 & 71.42 \\
 & ABBA & 72.93 & 62.29 & 79.00 & 74.74 & 80.99 & 80.00 & 74.75 & 65.37 & \textbf{73.76} \\
\rowcolor{cyan!8}
\cellcolor{white} & BoHA & 73.33 & 62.34 & 79.89 & 73.59 & 79.82 & 80.59 & 73.83 & 65.55 & 73.62 \\
\bottomrule
\end{tabular}
\end{table}

Table~\ref{tab:block_count_sweep_3b} gives the block-count sweep
referenced in Section~\ref{sec:rb_choice}. The sweep keeps the
LoRA-equivalent total rank fixed and varies the number of blocks. The
$b{=}8$ row is the canonical main-result setting.

\begin{table}[H]
\centering
\caption{Block-count sweep on Llama-3.2-3B commonsense reasoning at
fixed total rank $r_{\mathrm{tot}}=32$.}
\label{tab:block_count_sweep_3b}
\small
\begin{tabular}{rrrc}
\toprule
\textbf{$b$} & \textbf{$r_b$} & \textbf{$r_{\mathrm{tot}}$} & \textbf{Avg. Acc. (\%)} \\
\midrule
$2$ & $16$ & $32$ & $83.50$ \\
$4$ & $8$  & $32$ & $83.98$ \\
$8$ & $4$  & $32$ & $84.49$ \\
\bottomrule
\end{tabular}
\end{table}

\section{Equivalence at $b{=}1$}
\label{app:boundary_cases}
\label{sec:boundary_cases}
\label{sec:boundary}

Setting $b{=}1$ collapses the BoHA partition to a single block and
recovers the HiRA parameterization. We evaluate this equivalence with
a paired Llama-3.2-1B comparison. The BoHA--HiRA average-accuracy gap is
$+0.05$ percentage points. The associated paired $t$-test gives
$p=0.8868$, placing the residual within the bf16 and
Kaiming-initialization seed variability of this paired run. This agreement is consistent with
the algebraic reduction and separates the
$b{=}1$ endpoint from the $b>1$ blockwise settings studied in the
block-count and rank-budget controls.

\section{Additional Results on Arithmetic Reasoning}
\label{app:pertask}

We evaluate BoHA on a 40K-sample subset of MetaMathQA at Mistral-7B
and Gemma-2-9B, doubling the data budget of the main 20K
comparison. Results are reported in
Table~\ref{tab:appendix_pertask_arith40k}. BoHA exceeds the FFT
reference from ABBA on both Mistral-7B metrics and on
Gemma-2-9B GSM8K. Among the reported PEFT rows, BoHA leads on three of
four metric cells, with HiRA leading on Gemma-2-9B MATH at $40.02\%$
versus BoHA's $38.58\%$.

\begin{table}[H]
\centering
\caption{Results (\%) on Mistral-7B and Gemma-2-9B
across arithmetic reasoning benchmarks (GSM8K and MATH) on the
MetaMathQA-40K subset. \textsuperscript{\textdagger} indicates
results taken from ABBA~\citep{singhal2026abba}.}
\label{tab:appendix_pertask_arith40k}
\small
\setlength{\tabcolsep}{3pt}
\begin{tabular}{lrccrcc}
\toprule
\multirow{2}{*}{\textbf{Method}}
 & \multicolumn{3}{c}{\textbf{Mistral-7B}} & \multicolumn{3}{c}{\textbf{Gemma-2-9B}} \\
\cmidrule(lr){2-4}\cmidrule(lr){5-7}
 & \textbf{\#Params} & \textbf{GSM8K ($\uparrow$)} & \textbf{MATH ($\uparrow$)} & \textbf{\#Params} & \textbf{GSM8K ($\uparrow$)} & \textbf{MATH ($\uparrow$)} \\
\midrule
FFT\textsuperscript{\textdagger} & 7.24B   & 66.28             & 18.34             & 9.24B   & 79.89             & 39.44             \\
LoRA                             & 83.89M  & 64.44             & 16.92             & 108.04M & 76.95             & 34.84             \\
ABBA                             & 83.89M  & 65.35             & 18.16             & 108.04M & 79.83             & \underline{38.82} \\
HiRA                             & 83.89M  & \underline{66.19} & \underline{18.76} & 108.04M & \underline{79.91} & \textbf{40.02}    \\
\rowcolor{cyan!8}
BoHA                             & 83.89M  & \textbf{66.87}    & \textbf{19.98}    & 108.04M & \textbf{80.44}    & 38.58             \\
\bottomrule
\end{tabular}
\end{table}

\section{Reverse-direction continual learning diagnostic}
\label{app:cross_scale}

\begin{table}[H]
\centering
\caption{Reverse-direction continual-learning diagnostic
(arithmetic\,$\to$\,commonsense) on Llama-3.2-1B and Gemma-2-9B.
Reverse cells are single-seed runs that serve as a direction and
scale check rather than a fully replicated comparison.}
\label{tab:reverse_cl}
\small
\setlength{\tabcolsep}{5pt}
\begin{tabular}{llccccc}
\toprule
\textbf{Model} & \textbf{Method} & $R_{1,1}$ (\%, $\uparrow$) & $R_{2,1}$ (\%, $\uparrow$) & $R_{2,2}$ (\%, $\uparrow$) & BWT (\%, $\uparrow$) & ACC (\%, $\uparrow$) \\
\midrule
\multirow{4}{*}{Llama-3.2-1B}
 & LoRA  & \underline{16.63} & 3.12              & \underline{63.53} & $-13.51$             & \underline{33.33} \\
 & ABBA  & 14.13             & \underline{3.49}  & 60.20             & \textbf{$-10.64$}    & 31.85             \\
 & HiRA  & 15.50             & \textbf{3.70}     & 59.58             & \underline{$-11.80$} & 31.64             \\
\rowcolor{cyan!8}
\cellcolor{white} & BoHA  & \textbf{17.10}    & 3.40              & \textbf{65.05}    & $-13.70$             & \textbf{34.23}    \\
\midrule
\multirow{4}{*}{Gemma-2-9B}
 & LoRA  & 55.87             & 10.46             & 86.97             & $-45.41$             & 48.72             \\
 & ABBA  & 58.83             & 18.37             & \underline{88.48} & $-40.46$             & 53.42             \\
 & HiRA  & \underline{58.85} & \textbf{27.54}    & 87.84             & \textbf{$-31.31$}    & \textbf{57.69}    \\
\rowcolor{cyan!8}
\cellcolor{white} & BoHA  & \textbf{59.09}    & \underline{21.85} & \textbf{88.97}    & \underline{$-37.24$} & \underline{55.41} \\
\bottomrule
\end{tabular}
\end{table}

Table~\ref{tab:reverse_cl} reports the reverse-direction
continual-learning diagnostic for Llama-3.2-1B and Gemma-2-9B as a
direction and scale check on the forward 3B finding. At Gemma-2-9B
reverse, the family ordering replicates the forward 3B grouping:
HiRA and BoHA reach the highest stage-1 retention at $R_{2,1}$ of
$27.54\%$ and $21.85\%$, ABBA sits between the coupled pair and
LoRA at $18.37\%$, and LoRA falls last at $10.46\%$. The
cross-family ordering between $W_0$-coupled and $W_0$-free PEFT
updates is consistent with the 3B forward grouping. At
Llama-3.2-1B reverse, all four methods compress into an $R_{2,1}$
band between $3\%$ and $4\%$ where the gaps fall within evaluation
noise, so we treat the 1B reverse ranking as inconclusive. Within
the $W_0$-coupled family the BoHA versus HiRA order varies with
scale and direction: BoHA leads in 3B forward, HiRA leads BoHA by
$5.69\%$ in 9B reverse, and HiRA leads BoHA in 1B reverse. The
within-family ordering is therefore scale-dependent, while the
cross-family separation between $W_0$-coupled and $W_0$-free PEFT
updates is the signal emphasized in the main text.

\section{Code Generation Evaluation}
\label{app:code_gen}

As shown in Table~\ref{tab:code_gen}, BoHA achieves the best
pass@1 on three of four metric cells, leading at Llama-3.2-1B on
both HumanEval and HumanEval+ and at Llama-3.1-8B on HumanEval.
These results suggest that BoHA's gains are not confined to
reasoning. The two-model scope follows
HiRA~\citep{huang2025hira} and ABBA~\citep{singhal2026abba}, which
report code generation on the same two models.

\begin{table}[H]
\centering
\caption{Code-generation pass@1 (\%) on HumanEval and its
EvalPlus-augmented variant HumanEval+ at Llama-3.2-1B and
Llama-3.1-8B.}
\label{tab:code_gen}
\setlength{\tabcolsep}{6pt}
\begin{tabular}{lcccc}
\toprule
 & \multicolumn{2}{c}{\textbf{Llama-3.2-1B}}
 & \multicolumn{2}{c}{\textbf{Llama-3.1-8B}} \\
\cmidrule(lr){2-3}\cmidrule(lr){4-5}
\textbf{Method} & \textbf{HumanEval ($\uparrow$)} & \textbf{HumanEval+ ($\uparrow$)} & \textbf{HumanEval ($\uparrow$)} & \textbf{HumanEval+ ($\uparrow$)} \\
\midrule
LoRA & \underline{25.6} & 22.0             & \underline{56.7} & \textbf{53.7}    \\
ABBA & 25.0             & \underline{22.6} & 54.9             & 50.0             \\
HiRA & 25.0             & \underline{22.6} & 56.1             & 51.8             \\
\rowcolor{cyan!8}
BoHA & \textbf{26.2}    & \textbf{24.4}    & \textbf{57.3}    & \underline{52.4} \\
\bottomrule
\end{tabular}
\end{table}

\section{Frobenius Energy of Trained Updates}
\label{app:sv_extra}

We report the unnormalised Frobenius energy $\|\Delta W\|_F^2$ of
trained updates as a companion diagnostic to the effective rank and
$1\%$-threshold counts of Section~\ref{sec:spectral}. Both rank
diagnostics there cluster methods by $W_0$-coupling and structural
family. The energy diagnostic decouples from those orderings and is
dominated by adapter scaling conventions rather than by spectral
breadth.

\begin{figure}[H]
\centering
\includegraphics[width=\linewidth,height=0.20\textheight,keepaspectratio]{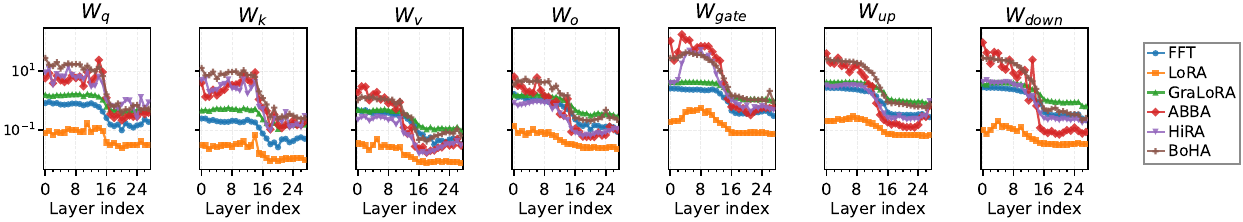}
\caption{Layer-wise sum of squared singular values for FFT, LoRA,
GraLoRA, ABBA, HiRA, and BoHA at Llama-3.2-3B on a log-y axis.}
\label{fig:sv_energy}
\end{figure}

\autoref{fig:sv_energy} reports this diagnostic layer by layer on
Llama-3.2-3B. The mean ordering on $\|\Delta W\|_F^2$ is
ABBA $>$ BoHA $>$ HiRA $>$ GraLoRA $>$ FFT $>$ LoRA across the seven
trained projections. The curves therefore follow a different ordering
from the effective-rank diagnostics in \autoref{fig:effective_rank}
and \autoref{fig:sv_counts}. All multiplicatively-scaled adapters sit
above FFT in absolute energy because their parameterisations include
$\alpha$-based scaling factors not present in the FFT update. ABBA has
the highest energy mean, but its effective rank in
\autoref{fig:effective_rank} remains mid-range at approximately $164$,
showing that unnormalised energy alone does not predict spectral
breadth. BoHA's
and HiRA's proximity to FFT in effective rank is consistent with the
$W_0$-coupled design.

\end{document}